\pdfoutput=1
\documentclass{article}
\usepackage{ijcai17}
\usepackage{graphicx}
\usepackage{subfigure}
\usepackage{times}
\usepackage{amsmath}
\usepackage{multirow}
\usepackage{color}
\usepackage{bm}
\usepackage{url}
\usepackage{calrsfs}

\title{Person Re-Identification by Deep Joint Learning of Multi-Loss Classification}
\author{Wei Li, Xiatian Zhu, Shaogang Gong\\ 
	Queen Mary University of London, London, UK  \\
	\{w.li, xiatian.zhu, s.gong\}@qmul.ac.uk}

\begin{document}

\maketitle

\begin{abstract}
	Existing person re-identification (re-id) methods 
	rely mostly on either localised or global feature
	representation {\em alone}. This ignores their joint benefit
	and mutual complementary effects. In this work, we show the
	advantages of jointly learning local and global features
	in a Convolutional Neural Network (CNN) by aiming
	to discover correlated local and global features in different context. Specifically, we
	formulate a method for joint learning of local and global
	feature selection losses designed to optimise person re-id when
	using {\em only} generic matching metrics such as the L2 distance.
	We design a novel CNN architecture for Jointly Learning
	Multi-Loss (JLML) of local and global discriminative feature optimisation subject
	concurrently to the same re-id labelled information. Extensive
	comparative evaluations demonstrate the advantages of this new
	JLML model for person re-id over a wide range of
	state-of-the-art re-id methods on five benchmarks (VIPeR, GRID,
	CUHK01, CUHK03, Market-1501). 
%In this paper, we propose an customized deep spatial constrained model (DSCM) to 
%address the person re-identification task under different camera views. The proposed DSCM captures the stable geometric structural information on pedestrians by learning a global-local image representation. Experimental results on several benchmark datasets demonstrate the effectiveness of our approach against the state-of-the-art approaches. 
\end{abstract}

\section{Introduction}

Person re-identification (re-id) is about matching identity classes in
detected person bounding box images from non-overlapping camera views
over distributed open spaces.
This is an inherently challenging task because person visual
appearance may change dramatically in different camera views from
different locations due to unknown changes in human pose,
illumination, occlusion, and background clutter \cite{gong2014person}.
Existing person re-id studies typically focus on either feature representation
\cite{ELF_ECCV08,SDALF_CVPR10,colorInvariants,SalienceReId_CVPR13,liao2015person,GOG,ma2017person} or 
matching distance metrics
\cite{KISSME_CVPR12,xiong2014person,PRDC,wang2014person,Anton_2015_CoRR,zhang2016learning,wang2016highly,wang2016human,wang2016pami,chen2017person}
or their combination in deep learning framework
\cite{li2014deepreid,ahmed2015improved,wangjoint,xiao2016learning,subramaniam2016deep,chen2016multi}.
Regardless, the overall objective is to 
obtain a view- and location-invariant (cross-domain) representation. We consider
that learning any matching distance metric is intrinsically learning a
global feature transformation across domains (two disjoint camera
views) therefore obtaining a ``normalised'' feature representation for
matching. 
% 
% Saliency based methods \cite{SalienceReId_CVPR13,hanxiao2014GTS}

%To seek robustness against person pose variance and appearance diversity, 
Most re-id features  
% either hand-engineered or deep learned, 
are typically hand-crafted to 
encode {\em local} topological and/or spatial structural information,
by different image decomposition schemes such as 
horizontal stripes \cite{ELF_ECCV08,colorInvariants},  
body parts \cite{SDALF_CVPR10}, 
and patches \cite{SalienceReId_CVPR13,GOG,liao2015person}. 
These localised features are effective 
for mitigating the person pose and detection misalignment in re-id matching. 
More recent deep re-id models \cite{xiao2016learning,wangjoint,chen2016multi,ahmed2015improved} 
benefit from the availability of larger scale datasets such as
CUHK03 \cite{li2014deepreid} and Market-1501 \cite{zheng2015scalable}
and from lessons learned on other vision
tasks \cite{krizhevsky2012imagenet,girshick2014rich}. 
In contrast to {\em local} hand-crafted features,
deep models, in particular Convolutional Neural Networks
(CNN) \cite{lecun1998gradient}, favour intrinsically in learning {\em
global} feature representations with a few exceptions. They have been
shown to be effective for re-id. 
%({\bf i added ``with a few
%exceptions'' due to that we did review two recent models that
%exploited both local and global feature learning})
%thanks to the built-in capability of extracting specific-to-abstract hierarchical
%visual pattens directly from raw pixels without the need for explicit image division.
% 
%The success of deep global re-id features is mainly due to
%the availability of large quantities of labelled identity classes
%in addition to the strong learning capacity of deep models. % with millions of parameters.

We consider that either local or global feature learning {\em alone} is
suboptimal. This is motivated by the human visual system
that leverages both global (contextual) and local (saliency) information concurrently
\cite{navon1977forest,torralba2006contextual}.
This intuition for {\em joint learning} aims to extract correlated
complementary information in different context whilst {\em satisfying the
same learning constraint}\footnote{In person re-id context, the learning constraint
refers to the image person identity label supervision.} 
therefore achieving more reliable recognition.
To that end, we need to address a number of non-trivial problems:
(i) the model learning behaviour in satisfying the same label
constraint may be different at the local and global levels;
(ii) any complementary correlation between local and global features
is unknown and may vary among individual instances, therefore must
be learned and optimised consistently across data; 
(iii) People's appearance in public scenes is diverse in both pattens
and configurations. This makes it challenging to learn correlations between local and
global features {\em for all appearances}.

This work aims to formulate a deep learning model for jointly
optimising local and global feature selections concurrently and 
to improve person re-id using {\em only} generic matching metrics such
as the L2 distance. We explore a deep learning approach for its
potential superiority in learning from large scale data
\cite{xiao2016learning,chen2016multi}.
For the bounding box image based person re-id, 
we consider the entire person in the bounding box as a {\em global scene context} and body
parts of the person as {\em local information sources}, both are subject to
the surrounding background clutter within a bounding box, and potentially
also misalignment and partial occlusion from bounding box detection.
In this setting, we wish to discover and optimise
jointly correlated complementary feature selections in the local and global
representations, both subject to the same label constraint concurrently. Whilst the
former aims to address pose/detection misalignment and occlusion by localised
fine-grained saliency information, the latter exploits holistic
coarse-grained context for more robust global matching.

To that end, we formulate a deep two-branch CNN architecture,
with one branch for learning localised feature selection (local branch)
and the other for learning global feature selection (global branch).
Importantly, the two branches are not independent but
synergistically correlated and jointly learned concurrently. 
This is achieved by: (i) imposing inter-branch interaction between the
local and global branches, and (ii) enforcing a separate learning
objective loss function to each branch
for learning independent discriminative capabilities, whilst being
subject to the same class label constraint.
Under such balancing between interaction and independence, 
we allow both branches to be learned concurrently 
for maximising their joint optimal extraction and selection of different
discriminative features for person re-id. 
We call this model the {\bf Joint Learning Multi-Loss} (JLML) CNN model.
To minimise poor learning due to inherent noise
and potential covariance, we introduce a structured feature selective
and discriminative learning mechanism into both the local and global
branches subject to a joint sparsity regularisation.

The {\bf contributions} of this work are:
{\bf (I)} We propose the idea of learning concurrently both local and global
feature selections for optimising feature discriminative capabilities
in different context whilst performing the same person re-id tasks. This is
currently under-studied in the person re-id literature to our best knowledge. 
{\bf (II)} 
We formulate a novel {\em Joint Learning Multi-Loss} (JLML) CNN model
for not only learning both global and local discriminative features in
different context by optimising multiple classification losses
on the same person label information concurrently, but also utilising their
complementary advantages jointly in coping with local misalignment and optimising
holistic matching criteria for person re-id.
{\bf (III)} 
We introduce a structured sparsity based feature selection learning mechanism
for improving multi-loss joint feature learning robustness w.r.t. noise and data
covariance between local and global representations.
Extensive comparative evaluations demonstrate the superiority of the
proposed JLML model over a wide range of existing state-of-the-art
re-id models on five benchmark datasets VIPeR \cite{ELF_ECCV08}, GRID \cite{loy2009multi},
CUHK01~\cite{li2012human}, CUHK03~\cite{li2014deepreid}, and Market-1501~\cite{zheng2015scalable}.
%

% \section{Related Work}

%\vspace{0.1cm}
%\noindent {\bf Related Works.} 
\section{Related Works}
The proposed JLML model considers
learning both local and global feature selections jointly for optimising
their correlated complementary advantages. This goes beyond existing methods
mostly relying on only one level of feature representation. 
Specifically, the JLML method is related to the saliency learning
based models \cite{SalienceReId_CVPR13,WangEtAl_BMVC14} in terms of 
modelling localised part importance.
However, these existing methods consider only the patch appearance statistics within individual
locations but no global feature representation learning, 
%({\bf check
%that Hanxiao's topic model also does not consider global features - i
%cannot remember now})
let alone the correlation and complementary information discovery
between local and global features as modelled by the JLML.

Whilst the more recent Spatially Constrained Similarity (SCS) model 
\cite{chen2016similarity}
and Multi-Channel Parts (MCP) network \cite{Cheng_TCP} consider both levels
of representation, the JLML model differs significantly from them:
%is particularly characterised by
%the superior learning capability for discovering 
%the underlying complementary information between the two levels of representation.
%More specifically, 
{\bf (i)} The SCS method focuses on supervised metric learning, 
whilst the JLML aims at joint discriminative feature learning 
and needs only generic metrics for re-id matching.
Also, hand-crafted local and global features are extracted {\em separately} in SCS 
without any inter-feature interaction and correlation learning involved,
as opposite to the joint learning of global and local feature selections concurrently
subject to the same supervision information in the JLML; 
%given local and global feature extraction and metric learning are separated;
%the complementary 
%is
% differs significantly in three aspects: 
%(i) The global-local feature interaction is ignored
%in \cite{chen2016similarity}, i.e. it is not jointly learned;
{\bf (ii)} The local and global branches of the MCP model are supervised and 
optimised by a triplet ranking loss, in contrast to
the proposed multiple classification loss design (Sec. \ref{sec:JLML_model}). 
%i.e. they are probably not sufficiently independent 
%but over correlated. 
Critically, this one-loss model learning is likely to impose negative influence on the discriminative 
feature learning behaviour for both branches
due to potential over-low pre-branch independence and over-high inter-branch correlation.
This may lead to suboptimal joint learning of local and global feature selections in model optimisation, 
as suggested by our evaluation in Section \ref{sec:ablation}.
%The complementary effect between the two levels of representation 
%is not effectively modelled and explored 
%due to the separate extraction of the two;
%({\bf need more explicit / specific - otherwise this is just a statement / claim
%by us without supporting evidence})
{\bf (iii)} In addition, the JLML is capable of performing structured feature sparsity
regularisation along with the multi-loss joint learning of local and global feature
selections for providing additional benefits (Sec. \ref{sec:ablation}).
Whilst similar in theory to the sparsity constraint on the supervised SCS metric learning,
we perform differently sparse generic feature learning 
{\em without} the need for supervised metric optimisation.

In terms of loss function, the HER model \cite{wang2016highly} 
similarly does not exploit pair-wise re-id labels but defines
a single identity label per training person for 
{\em regression loss} (vs. the classification loss in the JLML)
based re-id feature embedding optimisation.
Importantly, HER relies on the pre-defined feature (mostly hand-crafted local feature) 
{\em without} the capability of jointly learning global and local feature
representations and discovering their correlated complementary advantages
as specifically designed in JLML. 
%{\color{red} (May remove HER if no space in content or reference paper).}
% 
Also, the DGD \cite{xiao2016learning} model uses
the classification loss for model optimisation. 
However, this model considers only the global feature representation learning
of {\em one-loss} classification 
% and focuses on {\em domain-specific} knowledge mining,
as opposite to the proposed joint global and local feature learning of {\em multi-loss} classification
%for optimising {\em domain-generic} discriminative representation 
concurrently subject to maximising the same person identity matching.

%%% Hanxiao's BMVC %%%

% ({\bf again, this doesn't say anything meaningful. why
%this is useful?? our experimental results support its advantages?
%refer to specific comparative evaluations})

%=====================================
\section{Model Design}

\subsection{Problem Definition}

We assume a set of $n$ training images 
$\mathcal{I} = \{ \bm{I}_i \}_{i=1}^n$ with the corresponding 
identity labels as $\mathcal{Y} = \{y_i\}_{i=1}^n$.
These training images capture the visual appearance of 
$n_\text{id}$ (where $y_i \in [1,\cdots,
n_\text{id}]$) different people 
under non-overlapping camera views.
%with possibly significant viewing condition disparity.
We formulate a Joint Learning Multi-Loss (JLML) CNN model that 
aims to discover and capture concurrently
complementary discriminative information about a person image from
both local and global visual features of the image in order to
optimise person re-id under significant viewing 
condition changes across locations. 
This is in contrast to most existing re-id methods typically depending
only on either local or global features alone.

\subsection{Joint Learning Multi-Loss}
\label{sec:JLML_model}

\begin{figure} [h]
	\centering
	\includegraphics[width=1.00\linewidth]{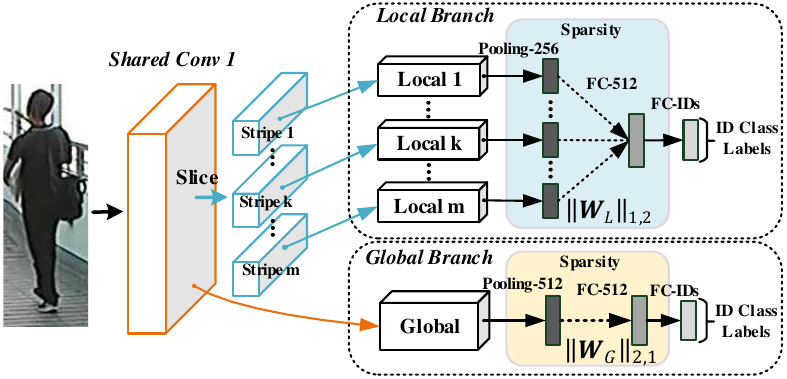}
	\vskip -0.1cm
	\caption{\footnotesize
		The Joint Learning Multi-Loss (JLML) CNN model architecture. 
		% ({\color{red} TODO: need to revise, the number $n$ is not consistent with later $M$.}) %The input image size of this network is $224\times224\times3$. 
	}
	\label{fig:pipline}
\end{figure}

% \noindent {\bf Network Architecture. }
The overall design of the proposed JLML model is depicted in
Figure \ref{fig:pipline}. This JLML model consists of a two-branches CNN network:
(1) One {\em local branch} of $m$ streams of an identical structure
with each stream learning the most discriminative local visual
features for one of $m$ local image regions of a person bounding box image;
%i.e. for {\em local appearance feature} modelling;
(2) Another {\em global branch} responsible for learning the most
discriminative global level features from the entire person image.
%i.e. for {\em global appearance feature} modelling.
% We aim to o
For concurrently optimising
per-branch discriminative feature representations and
discovering correlated complementary information between local and
global feature selections, a {\em joint learning} scheme that subjects
both local and global branches to the same identity label supervision
is considered with two underlying principles:

\vspace{0.1cm}
\noindent \textit{(I) Shared low-level features.}
We construct the global and local branches on a shared lower conv
layer, in particular the first conv layer\footnote{We found
empirically no clear benefits from increasing the number of shared
conv layers in our implementation.}, for facilitating inter-branch
common learning. 
The intuition is that, the lower conv layers capture low-level
features such as edges and corners which are common to all patterns in
the same images. This shared learning is similar in spirit to
multi-task learning \cite{argyriou2007multi}, where
the local and global feature learning branches are two related learning tasks.
Sharing the low-level conv layer reduces the model parameter size
therefore model overfitting risks. This is especially critical in
learning person re-id models when labelled training data is limited.
%We will evaluate the effects of the common ground for person re-id later
%(Section \ref{sec:ablation}).

\vspace{0.1cm}
\noindent \textit{(II) Multi-task independent learning subject to
shared label constraints.}
To maximise the learning of complementary discriminative features from
local and global representations, the remaining layers of the two
branches are learned independently subject to given identity labels. That
is, the JLML model aims to learn concurrently multiple identity
feature representations for different local image regions and the
entire image, all of which aim to maximise the {\em same} identity
matching {\em both} individually
and collectively at the same time. 
Independent multi-task learning aims to preserve both local saliency
in feature selection and global robustness in image
representation. 
To that end, the JLML model is designed to perform {\em multi-task
independent learning subject to shared identity label constraints} by
allocating each branch with a separate objective loss function. 
By doing so, the per-branch learning behaviour is conditioned
independently on 
the respective feature representation. We call this branch-wise loss
formulation as the {\bf MultiLoss} design. 
% except the shared conv layer (common ground),
%As a consequence, both branches can maximise their independent discriminative 
%learning capabilities from the same label supervision constraints in a {\em concurrent} manner
% This is made possible 
%with more learning freedom space allowed.
%In other words, such concurrent {\em independent branch evolution} is designed to provide
%more learning freedom space for optimising independent discriminative power
%of both local and global representations.
% but also and thus
%more likely {\em mutually-complementary} since each learning task focuses on 
%its own distinct characteristics.

\begin{table} [!h]
	\centering
	\footnotesize
	%\scalebox{0.8}{
	\renewcommand{\arraystretch}{1.3}
	\setlength{\tabcolsep}{0.07 cm}
	\vspace{-.5cm}
	\caption{\footnotesize
		JLML-ResNet39. 
		MP: Max-Pooling; AP: Average-Pooling; 
		S: Stride; SL: Slice; CA: Concatenation; G: Global; L: Local. 
	}
	\vskip 0pt %\vskip -6pt
	\begin{tabular}{|c|c||c|c|c|}
		\hline
		Layer \# & Layer & Output Size & Global Branch & Local Branch \\ \hline \hline
		1 & conv1      & $112\!\!\times\!\!112$  &\multicolumn{2}{c|}{$3\!\!\times\!\!3$, 32, S-2} \\ \hline 
		\multirow{4}{*}{9} & \multirow{4}{*}{conv2\_x}  & \multirow{3}{*}{G: $56\!\!\times\!\!56$}  &$3\!\!\times\!\!3$ MP, S-2 & SL-4, $2\!\!\times\!\!2$ MP, S-1 \\ \cline{4-5}
		& & \multirow{3}{*}{L: $28\!\!\times\!\!56$}   &  \multirow{3}{*}{$\begin{bmatrix} 1\!\!\times\!\!1, 32 \\  3\!\!\times\!\!3, 32 \\ 1\!\!\times\!\!1, 64 \\ \end{bmatrix} \!\!\times\!\! 3$} &  \multirow{3}{*}{$\begin{bmatrix} 1\!\!\times\!\!1, 16 \\  3\!\!\times\!\!3, 16 \\ 1\!\!\times\!\!1, 32 \\ \end{bmatrix} \!\!\times\!\! 3$}    \\ 
		& &   &    &      \\ 
		& &   &    &      \\ \hline
		\multirow{3}{*}{9}& \multirow{3}{*}{conv3\_x}  & \multirow{2}{*}{G: $28\!\!\times\!\!28$}    &  \multirow{3}{*}{$\begin{bmatrix} 1\!\!\times\!\!1, 64 \\  3\!\!\times\!\!3, 64 \\ 1\!\!\times\!\!1, 128 \\ \end{bmatrix} \!\!\times\!\! 3$} &  \multirow{3}{*}{$\begin{bmatrix} 1\!\!\times\!\!1, 32 \\  3\!\!\times\!\!3, 32 \\ 1\!\!\times\!\!1, 64 \\ \end{bmatrix} \!\!\times\!\! 3$}   \\
		& & \multirow{2}{*}{L: $14\!\!\times\!\!28$} & &\\
		& &   &    &      \\ \hline
		\multirow{3}{*}{9}& \multirow{3}{*}{conv4\_x}  & \multirow{2}{*}{G: $14\!\!\times\!\!14$}    &  \multirow{3}{*}{$\begin{bmatrix} 1\!\!\times\!\!1, 128 \\  3\!\!\times\!\!3, 128 \\ 1\!\!\times\!\!1, 256 \\ \end{bmatrix} \!\!\times\!\! 3$} &  \multirow{3}{*}{$\begin{bmatrix} 1\!\!\times\!\!1, 64 \\  3\!\!\times\!\!3, 64 \\ 1\!\!\times\!\!1, 128 \\ \end{bmatrix} \!\!\times\!\! 3$}   \\
		& & \multirow{2}{*}{L: $7\!\!\times\!\!14$} & & \\ 
		& &   &    &      \\ \hline
		\multirow{3}{*}{9}& \multirow{3}{*}{conv5\_x}  & \multirow{2}{*}{G: $7\!\!\times\!\!7$}    &  \multirow{3}{*}{$\begin{bmatrix} 1\!\!\times\!\!1, 256 \\  3\!\!\times\!\!3, 256 \\ 1\!\!\times\!\!1, 512 \\ \end{bmatrix} \!\!\times\!\! 3$} &  \multirow{3}{*}{$\begin{bmatrix} 1\!\!\times\!\!1, 128 \\  3\!\!\times\!\!3, 128 \\ 1\!\!\times\!\!1, 256 \\ \end{bmatrix} \!\!\times\!\! 3$}   \\
		& & \multirow{2}{*}{L: $4\!\!\times\!\!7$} & & \\ 
		& &   &    &      \\ \hline
		\multirow{2}{*}{1}& \multirow{2}{*}{fc} & \multirow{2}{*}{$1\!\!\times\!\!1$} & $7\!\!\times\!\!7$ AP  &  $4\!\!\times\!\!7$ AP, CA-4 \\ \cline{4-5}
		& &  & $\begin{bmatrix} 1\!\!\times\!\!1, 512 \end{bmatrix}$ & $\begin{bmatrix} 1\!\!\times\!\!1, 512 \end{bmatrix}$   \\ \hline
		
		1 & fc & $1\!\!\times\!\!1$    & ID\# & ID\# \\ \hline
	\end{tabular}%}
	\label{tab:model_arch}
	\vspace{-.3cm}
\end{table}

\vspace{0.1cm}
\noindent {\bf Network Construction.}
We adopt the Residual CNN unit \cite{he2016deep} as the JLML's building blocks 
due to its capacity for deeper model design whilst retaining a smaller
model parameter size\footnote{The choice of base network
is independent of our JLML model design. Other types, e.g. 
GoogLeNet \cite{szegedy2015going} or VGG-Net \cite{simonyan2014very},
can be readily applied in our model. 
%(see Table~\ref{tab:base_nets}).
}.
% ({\bf we need to fill in the missing parameter
%size in table 12})
%
Specifically, we customise the ResNet50 architecture in both layer and
filter numbers and design the JLML model as a 39 layers ResNet ({\bf
JLML-ResNet39}) tailored for re-id tasks. 
%and each local branch being as $50\%$ complex as the global one in conv layers. 
% and  with additional more layers found
% little performance gain but increasing model complexity and training cost. 
%
%We eliminate the Max-Pooling operation by applying a larger stride (i.e. 2 px) for
%increasing model efficiency.
% The non-linear ReLU activation is applied on each conv layer.
The configuration of JLML-ResNet39 is given in Table \ref{tab:model_arch}. 
Note that, the ReLU rectification non-linearity \cite{krizhevsky2012imagenet} 
after each conv layer is omitted for brevity.
%
%We call this ResNet units based model as {\bf JLML-ResNet39}.

\vspace{0.1cm}
\noindent {\bf Feature Selection.}
To optimise JLML model learning robustness against noise and diverse
data source, we introduce a feature selection capability in JLML by 
a structure sparsity induced
regularization \cite{kong2014exclusive,wang2013multi}. 
Our idea is to have a competing-to-survive mechanism in feature
learning that discourages irrelevant features whilst encourages
discriminative features concurrently in different local and
global context to maximise a shared identity matching objective.
To that end, we sparsify the global feature representation 
% upgrade the loss function (Eq \ref{eq:loss}) 
with a group LASSO \cite{wang2013multi}:
%$\ell_{2,1} = \| \bm{W}_G \|_{2,1} = 
%\sum_{i=1}^{c} \| \bm{w}_g^i \|_2$ %\cite{wang2013multi}, 
\begin{equation}
	\ell_{2,1} = \| \bm{W}_G \|_{2,1} = 
	\sum_{i=1}^{d_g} \| \bm{w}_g^i  \|_2
\end{equation}
%
%({\bf more details
%needed, be consistent with fig 2, need to show summation over L2})
%\begin{equation}
%%l_\text{global} = l + \lambda_\text{global} \| \bm{W}_G \|_{2,1}
%\ell_{2,1} = \| \bm{W}_G \|_{2,1}
%\label{eq:group_lasso4global}
%\end{equation}
%
where $\bm{W}_G = [\bm{w}_g^1, \cdots, \bm{w}_g^{d_g}] \in \mathcal{R}^{c_g \times d_g}$ is the parameter matrix of the global branch 
feature layer taking 
as input $d_g$ dimensional vectors from the previous layer
and outputting $c_g$ dimensional (512-D) feature representation.
% representation
% , that is, the last fc layer with 512 dimensions.
%
Specifically, with the $\ell_1$ norm applied on the $\ell_2$ norm of $\bm{w}_g^i$, 
our aim is to learn (tune) selectively feature dimension
importance subject to both the sparsity principle and the identity label
constraint simultaneously.

\begin{figure} [h!]
	\centering
	\includegraphics[width=1.01\linewidth]{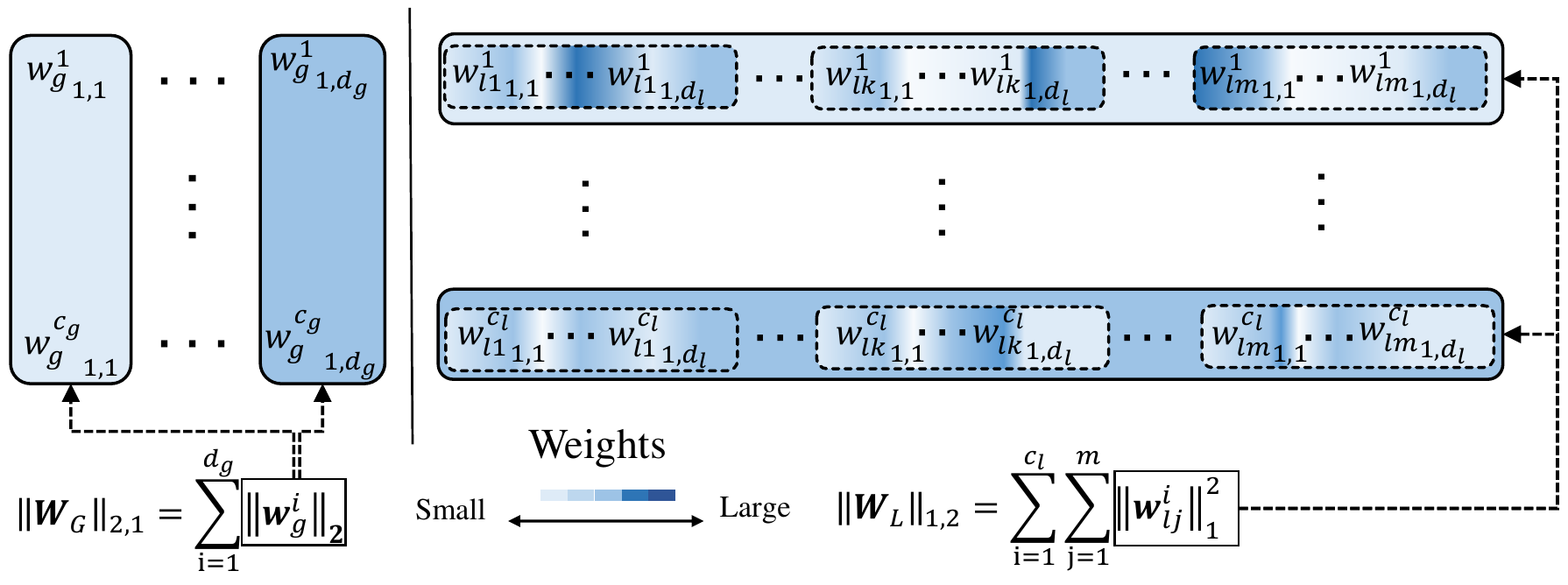}
	\vskip -0.3cm
	\caption{\footnotesize
		Group sparsity regularisations on fc layer parameter matrices
		($\bm{W}_G$ for the global branch and 
		$\bm{W}_L$ for the local branch)
		for selectively learning feature representations.
		%Elements in the matrices with darker colour illustrate larger values. 
		Solid and dashed rectangles denote $\ell_2$ norm and
		$\ell_1$ norm respectively. 
%		({\color{red} We may exclude this but for future use.})
	}
	\label{fig:lasso}
\end{figure}

% Similarly, we equip the local branch loss function with
Similarly, we also enforce a local feature sparsity constraint by 
an exclusive group LASSO \cite{kong2014exclusive}:
% $\ell_{1,2}
%= \| \bm{W}_L \|_{1,2} = \sum_{i=1}^{c)l} \sum_{j=1}^{d_l} \| \bm{w}_{l,j}^i \|_1^2$
\begin{equation}
	\ell_{1,2}
	= \| \bm{W}_L \|_{1,2} = 
	\sum_{i=1}^{c_l} \sum_{j=1}^{m} \| \bm{w}_{l,j}^i \|_1^2
\end{equation}
where
%({\bf this needs
%more details as how it's shown in fig 2, including the summation over
%L1. Also why local regularisation by L1 but global by L2? if possible,
%need a brief elaboration to justify})
%\begin{equation}
%% l_\text{local} = l + \lambda_\text{local} \| \bm{W}_L \|_{1,2}
%\ell_{1,2} = \| \bm{W}_L \|_{1,2}
%\label{eq:exc_group_lasso4local}
%\end{equation}
%where $\bm{W}_L = \[ \begin{array}{lcr}
%\bm{w}_{l,1}^1 & \cdots & \bm{w}_{l,d_l}^1 \\
%\cdots & \cdots & \cdots \\
%\bm{w}_{l,1}^c_l& \cdots & \bm{w}_{l,d_l^c_l} \end{array} \] $ 

\begin{equation}
	\bm{W}_L = 
	%\[ \begin{array} %{lcr}
	\begin{bmatrix}
		\bm{w}_{l,1}^{1\top} & \cdots & \bm{w}_{l,m}^{1\top} \\
		\cdots & \cdots & \cdots \\
		\bm{w}_{l,1}^{c_l\top}& \cdots & \bm{w}_{l,m}^{c_l\top} 
	\end{bmatrix}
	= \begin{bmatrix}
		\bm{w}_{l}^{1\top} \\
		\cdots \\
		\bm{w}_{l}^{c_l\top} 
	\end{bmatrix}
	%\end{array} \]
\end{equation}
is the parameter matrix of the local branch feature layer
with $m \times d_l$ and $c_l$ (512) as the input and output dimensions
($m$ the image stripe number).
The $\bm{w}_{l,j}^i \in \mathcal{R}^{d_l \times 1}$ defines
the parameter vector for contributing the $i$-th output feature dimension
from the $j$-th local input feature vector, $j \in [1, 2, \cdots, m]$. 
In particular, the $\ell_{2,1}$ regulariser performs sparse feature selection 
for individual image regions as below:
(1) We perform feature selective learning at the local region level
by enforcing the $\ell_1$ norm directly on $\bm{w}_{l,j}^i$,
conceptually similar to the group LASSO at the global level.
(2) We then apply a non-sparse smooth fusion with the $\ell_2$ norm
to combine the effects of different local features
weighted by the sparse $\bm{w}_{l,j}^i$. 
(3) Lastly, we exploit the $\ell_1$ norm again
at the level of $\bm{w}_l^{k}$ ($k \in [1,2,\cdots,c_l]$)
to learn the local 512-D feature representation selection.
%
%denotes the parameter matrix of the local branch fc
%feature layer with 512 dimensions.
%and $\lambda_\text{local}$ the trade-off co-efficient.
%
% and then smooth (non-sparse) fusion over all regions
%to merge regional features into a single local feature representation
% of 512-D. 
Figure \ref{fig:lasso} shows our structured sparsity
regularisations for both local and global feature selections.

\vspace{0.1cm}
\noindent {\bf Loss Function. }
% simipler, easy to prepare training by 
For model training,
we utilise the cross-entropy {\em classification} loss function
for both global and local branches so to optimise person {\em identity
classification} given training labels of multiple person classes
extracted from pair-wise labelled re-id dataset.
Formally, 
we predict the posterior probability $\tilde{y}_i$ of image $\bm{I}_i$
over the given identity label $y_i$:
\begin{equation}
{p}(\tilde{y}_{i} = y_i | \bm{I}_{i}) = \frac{\exp(\bm{w}_{y_i}^{\top} \bm{x}_{i})} {\sum_{k=1}^{|n_\text{id}|} \exp(\bm{w}_{k}^{\top} \bm{x}_{i})}
\label{eq:prob}
\end{equation}
where $\bm{x}_{i}$ refers to the feature vector of $\bm{I}_i$ from the corresponding branch,
and $\bm{W}_k$ the prediction function parameter
of training identity class $k$.
The training loss on a batch of $n_\text{bs}$ images is computed as:
% the average per-image loss:
\begin{equation}
l = - \frac{1}{n_\text{bs}}\sum_{i=1}^{n_\text{bs}}  \log \Big(p(\tilde{y}_{i} = y_i|\bm{I}_{i}) \Big)
\label{eq:loss_cls}
\end{equation}
Combined with the group sparsity based feature selection regularisations, 
we have the final loss function for the global and local branch
sub-networks as: 
\begin{equation}
l_\text{global} = l + \lambda_\text{global} \| \bm{W}_G \|_{2,1}, \;\; 
l_\text{local} = l + \lambda_\text{local} \| \bm{W}_L \|_{1,2}
\label{eq:loss}
\end{equation}
where $\lambda_\text{global}$ and $\lambda_\text{local}$
control the balance between the identity label loss and the feature
selection sparsity regularisation. 
We empirically set $\lambda_\text{local} = \lambda_\text{global} = 5\!\!\times\!\!10^{-4}$
by cross-validation in our evaluations. 
%and $\lambda_\text{global} = 5\!\!\times\!\!10^{-4}$ (Eq. \eqref{eq:loss}).
%({\bf how do we set them? what
%values? why? doesn't have to be cleaer but need be said})
%

\vspace{0.1cm}
\noindent {\bf Choice of Loss Function.}
Our JLML model learning deploys a {\em classification} loss
function. This differs significantly from the {\em
contrastive} loss functions used by most existing deep re-id methods
designed to exploit pairwise re-id labels defined by {\em both} positive and
negative pairs, such as the pairwise verification 
\cite{varior2016gated,subramaniam2016deep,ahmed2015improved,li2014deepreid},
triplet ranking \cite{Cheng_TCP},
or both \cite{wangjoint,chen2016multi}. Our JLML model training
does {\em not} use any labelled negative pairs inherent to all person
re-id training data, and we extract identity class labels from only
positive pairs.
The motivations for our JLML classification loss based learning are:
{\bf (i)} Significantly {\em simplified} training data batch
construction, e.g. random sampling with no notorious tricks required,
as shown by other deep classification methods \cite{krizhevsky2012imagenet}.
This makes our JLML model more scalable in real-world applications with
very large training population sizes when available. 
This also eliminates the {\em undesirable} need for 
carefully forming pairs and/or triplets in preparing re-id training
splits, as in most existing methods, due to the inherent imbalanced
negative and positive pair size distributions. 
{\bf (ii)} Visual psychophysical findings suggest that 
representations optimised for classification tasks generalise well to
novel categories \cite{edelman1998representation}. We consider that
re-id tasks are about model generalisation to unseen test identity classes
given training data on {\em independent} seen identity classes. 
Our JLML model learning exploits this general classification learning
principle beyond the strict pair-wise relative verification loss in
existing re-id models.

\subsection{Model Training}
We adopt the standard Stochastic Gradient Descent (SGD) optimisation algorithm \cite{krizhevsky2012imagenet}
to perform the batch-wise joint learning of local and
global branches.
Note that, with SGD we can naturally synchronise the optimisation processes 
of the two branches by constraining
their learning behaviours subject to the same identity label information
at each update.
This is likely to avoid representation learning divergence between two branches
and help enhance the correlated complementary learning capability.
%
%This synchronisation fits well the standard stochastic batch-wise optimisation algorithm
%(e.g. stochastic gradient descent)
%in deep model learning without the need for any notorious modification.

\subsection{Re-Id by Generic Distance Metrics}
%Once the JLML model is learned, we obtain a combined local-global deep
%feature representation as a 1,024 dimensional feature vector
% (joint fc layer in Table~\ref{tab:model_arch}). 
{Once the JLML model is learned, 
we obtain a 1,024-D joint representation
by concatenating the local (512-D) and global (512-D) feature vectors
(the fc layer in Table~\ref{tab:model_arch}).}
For person
re-id, we deploy this 1,024-D deep feature representation using {\em only}
a generic distance metric {\em without} camera-pair specific distance
metric learning, e.g. L2 distance.
Specifically, given a test probe image $\bm{I}^p$ from one camera view
and a set of test gallery images 
$\{\bm{I}_i^g\}$ from other non-overlapping camera views:
(1) We first compute their corresponding 1,024-D feature vectors by
forward-feeding the images to the trained JLML model, 
denoted as $\bm{x}^p=[\bm{x}_g^p; \bm{x}_l^p]$ and 
$\{\bm{x}_i^g=[\bm{x}_g^g; \bm{x}_l^g]\}$. 
(2) We then compute L2 normalisation on the global and local features,
separately. (3) Lastly we compute the cross-camera matching
distances between $\bm{x}^p$ and $\bm{x}_i^g$ by some generic matching
metric, e.g. L2 distance. We then rank all
gallery images in ascendant order by their L2 distances to the probe image.  
The probabilities of true matches of probe person images in Rank-1 and
among the higher ranks indicate the goodness of the learned
JLML deep features for person re-id tasks.

%=====================================
\section{Experiments}

%In this section, we compare our approach with the state-of-the-art approaches on several datasets. We report results of our proposed model on four challenging datasets with the evaluation criteria. In addition, the cross domain transfer learning on the trained networks with cross dataset evaluation also reported.

%\subsection{Training the Network}
%\vspace{0.1cm}

\begin{figure} [ht]
	\centering
	\subfigure[\scriptsize VIPeR]{
	\includegraphics[width=0.17\linewidth]{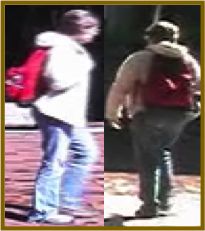}
	}
	\subfigure[\scriptsize GRID]{
		\includegraphics[width=0.17\linewidth]{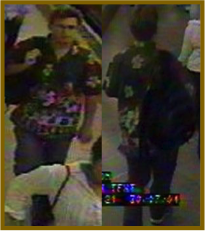}
	}
	\subfigure[\scriptsize CUHK01]{
		\includegraphics[width=0.17\linewidth]{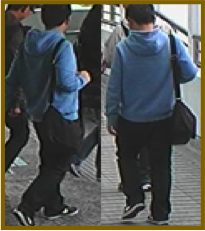}
	}
	\subfigure[\scriptsize CUHK03]{
		\includegraphics[width=0.17\linewidth]{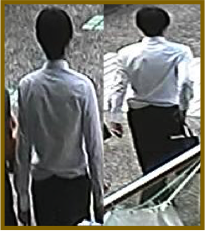}
	}
	\subfigure[\scriptsize Market]{
		\includegraphics[width=0.17\linewidth]{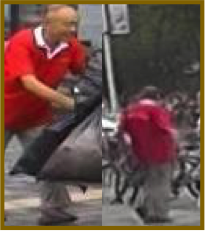}
	}
	\vskip -0.5cm
	\caption{\footnotesize
		Example cross-view image pairs from five re-id datasets.
		% with two images of one column corresponding to the same person.
		%({\color{red} TODO: need to revise})
	}
	\label{fig:dataset}
\end{figure}

%\vspace{0.1cm}
\noindent {\bf Datasets.}
For evaluation, we used five benchmarking re-id datasets,
VIPeR~\cite{ELF_ECCV08}, 
GRID~\cite{loy2009multi},
CUHK01~\cite{li2012human}, 
CUHK03~\cite{li2014deepreid}, and 
Market-1501~\cite{zheng2015scalable}.   
%
% TODO: add some words on dataset features.
%
Figure \ref{fig:dataset} shows some examples of person bounding box
images from these datasets.
The datasets are collected by different data sampling protocols
from different environments, where: 
{\bf (a)} VIPeR has one image per person per view,
with low-resolution under severe lighting change.
{\bf (b)} GRID provides one image per person per view, with additional
images for 775 distracting persons under very poor lighting from underground stations.
{\bf (c)} CUHK01 contains two images person per view from a university campus.
{\bf (d)} CUHK03 consists of up to five images per person per view,
obtained by both manually labelled and auto-detected person bounding
boxes with the latter posing a more challenging re-id task due to
detection bounding box misalignment and occlusion. 
{\bf (e)} Market-1501 has variable numbers of images per person per
view captured from a supermarket, with all bounding boxes
automatically detected. 
These datasets present a wide range of re-id evaluation scenarios with
different population sizes under different challenging viewing
conditions (Table \ref{tab:dataset_stats}).

%and our experimental settings with the training and testing splits.
\begin{table}[h!] \footnotesize
	\centering
	%\scalebox{0.8}{
	\renewcommand{\arraystretch}{1}
	\setlength{\tabcolsep}{0.02cm}
	\vspace{-0.5cm}
	\caption{\footnotesize
		Settings of person re-id datasets.
		TS: Test Setting;
		SS: Single-Shot; MS: Multi-Shot.
		SQ: Single-Query; MQ: Multi-Query.
	}
	\vskip 0pt %\vskip -6pt
	\begin{tabular}{|c||c|c|c|c|c|c|c|}
		\hline
		Dataset  & 
		{Cams} &
		{IDs} & 
		{Train IDs} & 
		{Test IDs} &
		{Labelled} & 
		{Detected} &
		TS \\ \hline \hline %
		VIPeR & 2 & 632 & 316 & 316 & 1,264 & 0 & SS \\  % \hline
		GRID & 8 & 250 & 125 & 125 & 1,275 & 0 & SS \\  % \hline
		CUHK01 & 2 & 971 & 871/485 & 100/486 & 1,942 & 0 & SS/MS\\
		CUHK03 & 6 & 1,467 & 1,367 & 100 & 14,097 & 14,097 & SS\\
		Market & 6  & 1,501& 751 & 750 & 0 & 32,668 & SQ/MQ \\
		\hline
	\end{tabular}%}
	\label{tab:dataset_stats}
	\vspace{-0.3cm}
\end{table}

\vspace{0.1cm}
\noindent {\bf Evaluation Protocol.}
We adopted the standard supervised re-id setting to evaluate the proposed JLML model
(Sec. \ref{sec:eval_specific}).
The training and test data splits and testing settings of each dataset is given in Table \ref{tab:dataset_stats}.
Specifically, on VIPeR, we split randomly the whole population (632 people) into two halves: One for training (316) and another for testing (316). 
We repeated $10$ trials of random people splits and used the averaged results.
On CUHK01, we considered two training/test splits:  
485/486 \cite{liao2015person}
and 871/100 \cite{ahmed2015improved}.
Again, we reported the results averaged over 10 random trials for either split.
On GRID, the training/test split were 125/125 with 775 distractor people included in the test gallery.
We used the benchmarking 10 people splits \cite{loy2009multi} and the averaged performance.
On CUHK03, following \cite{li2014deepreid} we repeated 20 times of
random 1260/100 training/test splits and reported the averaged
accuracies under the single-shot evaluation setting. 
On Market-1501, we used the standard training/test split (750/751) \cite{zheng2015scalable}.
We used the cumulative matching characteristic (CMC) to measure
re-id accuracy on all benchmarks, except on Market-1501 we also used
in addition the recall measure of multiple truth matches by mean
Average Precision (mAP), i.e. first computing the area under the
Precision-Recall curve for each probe, then calculating the mean of
Average Precision over all probes \cite{zheng2015scalable}. 

%Finally, we also considered a more scalable and realistic re-id test
%setting where the test data is from an entirely different domain
%(dataset) {\em without} any labelled information for model training,
%unlike the existing re-id settings where training/test splits are
%drawn from the same domain (dataset). Results are discussed in
%Sec. \ref{sec:eval_generic}.

\vspace{0.1cm}
\noindent {\bf Competitors.}
We compared the JLML model against $10$ existing
state-of-the-art methods as listed in Table \ref{tab:methods}.
They range from hand-crafted and deep learning features to
domain-specific distance metric learning methods. 
We summarise them into three categories:
{\bf (A)} Hand-crafted (feature) with domain-specific distance learning (metric);
{\bf (B)} Deep learning (feature) with domain-specific deep verification metric learning;
{\bf (C)} Deep learning (feature) with generic non-learning L2 distance (metric).
%The proposed JLML model belongs to the last category.
% In comparison, XXX
%including $4$ non-deep ({\em hand-crafted feature + learned metric}) methods: 
%(1) LOMO + XQDA~\cite{liao2015person}, 
%(2) GOG~\cite{matsukawa2016hierarchical} + XQDA, 
%(3) LOMO + NFST~\cite{zhang2016learning},
%(4) LOMO\&KCCA~\cite{lisanti2014matching} + NFST,
%(5) SCS \cite{chen2016similarity}
%
%and $7$ deep ({\em generic discriminative feature learning + non-learned metric}) alternatives: 
%DCNN+~\cite{ahmed2015improved}, 
%DGD~\cite{xiao2016learning},
%MCP~\cite{Cheng_TCP}, 
%SLSTM \cite{varior2016siamese}, 
%S-CNN \cite{varior2016gated}, 
%X-Corr~\cite{subramaniam2016deep},
%MTDnet \cite{chen2016multi}. 

%Colour Name (CN) \cite{van2007learning}

\begin{table} [h!]
	\centering
	\scriptsize
	%\footnotesize
	\renewcommand{\arraystretch}{1}
	\setlength{\tabcolsep}{0.025 cm}
	\vspace{-.5cm}
	\caption{\footnotesize
		Person re-id method categorisation by features and metrics.
		% Method name is indicated in bold.
		Cat: Category;
		DL: Deep Learning;
		CPSL: Camera-Pair Specific Learning;
		DVM: Deep Verification Metric;
		DVM,L2: Ensemble of DVM and L2;
		CHS: Fusion of Colour, HOG, SILPT features.
		% $^\text{HC}$: Hand-Crafted; $^\text{DL}$: Deep Learned.
	}
	\vskip 0pt %\vskip -6pt
	\begin{tabular}{|c||c|c|c|c|c|}
		\hline
		\multirow{2}{*}{Cat}
		& \multirow{2}{*}{Method} 
		& \multicolumn{2}{c|}{Feature} & \multicolumn{2}{c|}{Metric} \\ 
		\cline{3-6}
		& & Hand-Crafted & DL & CPSL & Generic \\ 
		\hline 
		\hline
		\multirow{4}{*}{A}
		& {\bf XQDA} \cite{liao2015person} 
		& LOMO & - 
		& XQDA & - \\   %\hline
		& {\bf GOG} \cite{matsukawa2016hierarchical} 
		& GOG & - 
		& XQDA & - \\  %\hline
		%
%		& \cite{zhang2016learning}
%		& LOMO & - 
%		& NSFT & - \\	
		%
		& {\bf NFST} \cite{zhang2016learning} 
		& LOMO, KCCA & - 
		& NSFT & - \\ 
		& {\bf SCS} \cite{chen2016similarity} 
		& CHS & -
		& SCS & - \\
		\hline
		%
%		B
%		& \cite{varior2016siamese}
%		& LOMO\&CN & SLSTM 
%		& - & L2 \\
%		\hline 
		% 
		\multirow{3}{*}{B}
		& {\bf DCNN+} \cite{ahmed2015improved} 
		& - & DCNN+ 
		& DVM & - \\
		& {\bf X-Corr} \cite{subramaniam2016deep} 
		& - & X-Corr 
		& DVM & - \\ 
%		& \cite{wangjoint}
%		& - & \bf SICI
%		& DVM\&L2 & - \\  
		& {\bf MTDnet} \cite{chen2016multi}
		& - & MTDnet 
		& DVM, L2 & - \\
		\hline
		\multirow{4}{*}{C}
		& {\bf S-CNN} \cite{varior2016gated} 
		& - & S-CNN 
		& - & L2 \\ 
		& {\bf DGD} \cite{xiao2016learning} 
		& - & DGD 
		& - & L2 \\ 
		& {\bf MCP} \cite{Cheng_TCP} 
		& - & MCP 
		& - & L2 \\
		%
		%\cline{2-5}
		& {\bf JLML} (Ours)
		& - & {JLML} 
		& - & L2 \\
		\hline
	\end{tabular}%}
	\label{tab:methods}
	\vspace{-.3cm}
\end{table}

\vspace{0.1cm}
\noindent {\bf Implementation.}
We used the Caffe framework~\cite{jia2014caffe} for our JLML model implementation. 
We started by pre-training the JLML model on ImageNet (ILSVRC2012).
Subsequently, for CUHK03 or Market, we used only their own training data 
for model fine-tuning, i.e. ImageNet $\rightarrow$ CUHK03/Market;
For CUHK01 or VIPeR or GRID, we pre-trained JLML on CUHK03+Market (whole datasets),
and then fine-tuned on their respective training images, i.e.
ImageNet $\rightarrow$ CUHK03+Market $\rightarrow$ CUHK01 / VIPeR / GRID.
All input person images were resized to $224\!\times\!224$ in pixel.
For local branch, according to a coarse body part layout 
we evenly decomposed the whole shared convolutional feature maps (i.e. the entire image) 
into four ($m=4$) horizontal strip-regions. 
%We empirically set $\lambda_\text{local} = 5\!\!\times\!\!10^{-4}$ %(Eq. \eqref{eq:exc_group_lasso4local})
%and $\lambda_\text{global} = 5\!\!\times\!\!10^{-4}$ (Eq. \eqref{eq:loss}).
%
%For model pre-training with ImageNet, 
%we set the learning rate 0.01, the momentum 0.9, the iteration number 50,000, batch size 32.
%For model training, we set the learning rate 0.01, momentum 0.9, iterations 50,000, batch size 32.
We used the same parameter settings (summarised in Table \ref{tab:param}) 
for pre-training and training the JLML model on all datasets.
We also adopted the stepped learning rate policy, 
e.g. dropping the learning rate by a factor of 10 every 100K iterations for JLML pre-training
and every 20K iterations for JLML training.
% ({\bf \color{red}TODO: Wei needs to add more details.})
We utilised the L2 distance as the default matching metric, unless stated otherwise.

\begin{table} [!h]
	\centering
	\footnotesize
	%\scalebox{0.8}{
	\renewcommand{\arraystretch}{1}
	\setlength{\tabcolsep}{0.2 cm}
	\vspace{-0.3cm}
	\caption{\footnotesize
		JLML training parameters. BLR: base learning rate; LRP: learning rate policy;
		MOT: momentum; IT: iteration; BS: batch size.
		%
		% {\color{red} TODO: need to add the strategy of learning rate decay. }
		%lr: learning rate; bz: batch size.
	}
	\vskip 0pt %\vskip -6pt
	\begin{tabular}{|c||c|c|c|c|c|}
		\hline
		Parameter 
		& BLR & LRP & MOT & IT \# & BS \\ \hline \hline
		Pre-train
		& 0.01 & step (0.1, 100K) & 0.9 & 300K & 32 \\ \hline
		Train
		& 0.01 & step (0.1, 20K) & 0.9 & 50K & 32 \\ 
		\hline
	\end{tabular}%}
	\label{tab:param}
	\vspace{-0.3cm}
\end{table}

%\nocite{li2012human}
%\nocite{jia2014caffe}
%\subsection{Comparisons to State-Of-The-Arts}
\subsection{Conventional Intra-Domain Re-Id Evaluations}
\label{sec:eval_specific}
We conducted extensively comparative evaluations on conventional % (intra-dataset) 
supervised learning based person re-id tasks.

\vspace{0.1cm}
\noindent {\bf (I) Evaluation on CUHK03. }
%We compare our deep spatial constrained model (DSCM) with several methods in recent years, including 
Table~\ref{tab:res_cuhk03} shows the comparisons of JLML against $8$
existing methods on CUHK03. 
It is evident that JLML outperforms existing methods in all categories
on both labelled and detected bounding boxes,
surpassing the 2nd best performers DGD and X-Corr on corresponding labelled and
detected images in Rank-1 by $7.9\%$(83.2-75.3) and $8.6\%$(80.6-72.0)
respectively. 
X-Corr/GOG/JLML also suffer the least from auto-detection
misalignment, indicating the robustness and competitiveness of the
joint learning approach to mining complementary local and global
discriminative features.

% and \ref{tab:res_market} illustrates the recognition performance comparison of these methods.

\begin{table} %[h]
	\centering
	% \scriptsize
	\footnotesize
	%\scalebox{0.8}{
	\renewcommand{\arraystretch}{1}
	\setlength{\tabcolsep}{0.1 cm}
	\vspace{-.3cm}
	\caption{\footnotesize
		CUHK03 evaluation. $1^\text{st}/2^\text{nd}$ best in
		red/blue.
		% $^\text{HC}$: Hand-Crafted; $^\text{DL}$: Deep Learned.
		% of state-of-the-art algorithms at rank 1, 5, 10 and 20 for the CUHK03 dataset.
	}
	\vskip 0pt %\vskip -6pt
	\begin{tabular}{|c||c|cccc|cccc|}
		\hline
		%		Dataset  & 
		%		\multicolumn{8}{c|}{CUHK03} \\ \hline % \citep{zheng2015scalable}
		\multirow{2}{*}{Cat}
		& Annotation &  \multicolumn{4}{c|}{Labelled} &\multicolumn{4}{c|}{Detected} \\ \cline{2-10}
		& Rank (\%) & R1 & R5 & R10 & R20 & R1 & R5 & R10 & R20 \\ \hline \hline
		\multirow{3}{*}{A}
		& XQDA & 55.2  & 77.1  &  86.8 &  83.1   & 46.3  & 78.9  &  83.5 &  93.2 \\   %\hline
		& GOG & 67.3  & 91.0  &  96.0 &  -   & 65.5  & 88.4  &  93.7 &  - \\  %\hline
		%
		%NSFT & 58.9 & 85.6 & 92.5 & 96.3	& 53.7 & 83.1 & 93.0 &  \color{blue}  94.8 \\	
		& NSFT & 62.5  & 90.0  & 94.8 &  98.1   & 54.7  & 84.7 &  \color{blue}  94.8 &  95.2 \\ 
		\hline
		\multirow{3}{*}{B}
		& DCNN+ & 54.7  & 86.5  &  93.9 &  98.1   & 44.9  & 76.0  &  83.5 &  93.2 \\
		& X-Corr & 72.4  & 95.5  &  - &  \color{blue}  98.4  &  \color{blue}  72.0  &  \color{blue}  96.0  &  - &  \color{blue}   98.2 \\ 
		% & SICI & -  & -  &  - &  - & 52.1 & 84.9 & 92.4 & - \\
		& MTDnet & 74.7 &  \color{blue}  96.0 &  \color{blue}   97.5 &  - &  -   & -  & -  &  - \\
		\hline
		% SLSTM & -  & -  &  - &  -   & 57.3  & 80.1  &  88.3 &  - \\ 
		\multirow{3}{*}{C}
		& S-CNN & -  & -  &  - &  -   & 68.1  & 88.1  &  94.6 &  -\\ 
		& DGD &  \color{blue}   75.3  & -  &  - &  - & -  & -  &  - & - \\ 
		% \hline
		%		{\bf JLML} & \textbf{81.0} & {95.8}  &  \textbf{98.7} &  \textbf{99.4}  & \textbf{80.5}  & 95.5  &  \textbf{97.7} &  \textbf{98.8} \\ 
		& {\bf JLML} & \color{red} 83.2 & \color{red} 98.0 & \color{red}  99.4 &  \color{red}  99.8
		&  \color{red}  80.6 &  \color{red}  96.9 &  \color{red}  98.7 &   \color{red}  99.2 \\
		\hline
	\end{tabular}%}
	\label{tab:res_cuhk03}
	\vspace{-.1cm}
\end{table}

\vspace{0.1cm}
\noindent {\bf (II) Evaluation on Market-1501. }
We evaluated JLML against four existing models on Market-1501.
Table \ref{tab:res_market} shows the clear performance superiority of JLML
over all state-of-the-arts with more significant Rank-1 advantages
over other methods compared to CUHK03, giving $19.3\%$(85.1-65.8) (SQ) and
$13.7\%$(89.7-76.0) (MQ) gains over the 2nd best S-CNN. %than on CUHK03 ($8.6\%$). 
This further validates the advantages of our joint learning of
multi-loss classification for optimising re-id especially when the
re-id test population size increases ($751$ people on Market-1501
vs. $100$ people on CUHK03).

\begin{table} [!h]
	\centering
	\footnotesize
	%\scalebox{0.8}{
	\renewcommand{\arraystretch}{1}
	\setlength{\tabcolsep}{0.3 cm}
	\vspace{-.5cm}
	\caption{\footnotesize
		% Performance comparisons on 
		Market-1501 evaluation. $1^\text{st}/2^\text{nd}$ best in red/blue.
		All person bounding box images were auto-detected.
	}
	\vskip 0pt %\vskip -6pt
	\begin{tabular}{|c||c|cc|cc|}
		\hline
		%		Dataset  & 
		%		\multicolumn{4}{c|}{Market-1501} \\ \hline % \citep{zheng2015scalable}
		\multirow{2}{*}{Cat}
		& Query Type &  \multicolumn{2}{c|}{Single-Query} &\multicolumn{2}{c|}{Multi-Query} \\ \cline{2-6}
		& Measure (\%)    
		& R1 & mAP & R1 & mAP  \\ \hline \hline
		\multirow{3}{*}{A}
		& XQDA &  43.8 &  22.2  &  54.1 &  28.4\\   %\hline
		& SCS &  51.9 &  26.3 &  - &  -  \\  %\hline
		%
		% DNS & 55.4 & 29.9 & 68.0 & 41.9 \\
		& NFST & 61.0 & 35.6  &  71.5 &  46.0  \\ 
		\hline
		%Siamese-LSTM  & - & -  &  61.6 &  35.3  \\ 
		\multirow{2}{*}{C}
		& S-CNN &  \color{blue}  65.8 &  \color{blue}  39.5  &  \color{blue}  76.0 &  \color{blue}  48.4  \\ 
		% DGD & & & & \\ 
		% \hline
		& {\bf JLML} &  \color{red} {85.1} &  \color{red} {65.5}  &   \color{red} {89.7} &   \color{red} {74.5}  \\ 
		\hline
	\end{tabular}%}
	\label{tab:res_market}
	\vspace{-.3cm}
\end{table}

\vspace{0.1cm}
\noindent {\bf (III) Evaluation on CUHK01.}
We compared our JLML model with $8$ state-of-the-art methods on CUHK01.
Table \ref{tab:res_cuhk01} shows that 
JLML surpasses clearly all compared models under both training/test splits in single- and multi-short settings.
Moreover, JLML outperforms in Rank-1 ($76.7\%$) the best hand-crafted
feature method NFST (R1 $69.1\%$) when the training population size is
small (486 people). % a general challenging situation for deep models.
When the training population size increases (871 people),
JLML is even more effective than all deep competitors in exploiting
extra training classes by inducing more identity-discriminative joint person features in distinct context.
For example, JLML gains $5.8\%$(87.0-81.2) more Rank-1 than the 2nd
best method X-Corr in single-shot re-id, further improved the gain of
$4.8\%$(69.8-65.0) under the 486/485 split.
These results show consistent superiority and robustness of the proposed JLML model
over the existing methods.
%in dealing with various numbers of training data for extracting 
%discriminative global and local appearance representation. 

\begin{table} [!h]
	\centering
	\footnotesize
	\renewcommand{\arraystretch}{1}
	\setlength{\tabcolsep}{0.1 cm}
	\vspace{-.3cm}
	\caption{\footnotesize
		CUHK01 evaluation. $1^\text{st}/2^\text{nd}$ best in bold/typewriter.
		% SS: Single-Shot; MS: Multi-Shot.
	}
	\vskip 0pt %\vskip -6pt
	\begin{tabular}{|c||c|cccc|cccc|}
		\hline
		%		Dataset  & 
		%		\multicolumn{8}{c|}{CUHK01} \\ \hline % \citep{zheng2015scalable}
		\multirow{2}{*}{Cat}
		& Split &  \multicolumn{4}{c|}{871/100 split} &\multicolumn{4}{c|}{486/485 split} \\ \cline{2-10}
		& Rank (\%) & R1 & R5 & R10 & R20 & R1 & R5 & R10 & R20 \\ \hline \hline
		&  \multicolumn{9}{c|}{Single-Shot Testing Setting} \\ \cline{1-10}
		\multirow{1}{*}{A}
		& GOG & -  & -  &  - &  - 
		& 57.8 & 79.1 & 86.2 & 92.1 \\
		\hline
		\multirow{3}{*}{B}
		& DCNN+ & 65.0  & -  &  - &  -   
		& 47.5  & 71.6  &  80.3 &  87.5 \\  %\hline		
		& X-Corr & \color{blue} 81.2  & \color{red} {97.3}  &  - & \color{blue} 98.6  
		 & 65.0  & \color{red} {89.7}  &  - & 94.4 \\ 
		& MTDnet & 78.5 & 96.5 & \color{blue} 97.5 & - & -  & -  &  - &  - \\ \hline
		\multirow{3}{*}{C}
		%\hline
		& DGD & -  & -  &  - &  -   
		& \color{blue} 66.6  & -  &  - &  -\\ 
		& MCP & -  & -  &  - &  -   
		& 53.7  & 84.3  & \color{blue} 91.0 & \color{red} 96.3 \\
		% \hline
		& {\bf JLML} & \color{red} {87.0} & \color{blue} 97.2  & \color{red} {98.6} & \color{red} {99.4}  
		& \color{red} {69.8 } & \color{blue} 88.4  & \color{red} {93.3} & \color{red} 96.3 \\ %single-shot

		\hline \hline
		&  \multicolumn{9}{c|}{Multi-Shot Testing Setting} \\ \cline{1-10}
		\multirow{3}{*}{A}
		& XQDA & -  & -  &  - &  -   & 63.2  & 83.9  &  90.0 &  94.2 \\   %\hline
		& GOG & -  & -  &  - & - & 67.3  & \color{blue} 86.9  & \color{blue} 91.8 & \color{blue} 95.9 \\  %\hline
		
		& NFST		& -  & -  & - & -   & \color{blue} 69.1  & \color{blue} 86.9 & \color{blue} 91.8 &  95.4 \\ 
		\hline
		\multirow{1}{*}{C}
		& {\bf JLML} & \color{red} 91.2 & \color{red} 98.4 & \color{red} 99.2 & \color{red} 99.8
		& \color{red} 76.7 & \color{red} 92.6 & \color{red} 95.6 & \color{red} 98.1 \\ % multi-shot
		\hline
	\end{tabular}%}
	\label{tab:res_cuhk01}
	\vspace{-.3cm}
\end{table}

\vspace{0.1cm}
\noindent {\bf (IV) Evaluation on VIPeR. }
We evaluated the performance of JLML against $8$ strong competitors on VIPeR,
a more challenging test scenario with fewer training
classes ($316$ people) and lower image resolution.
On this dataset, the best performers are 
hand-crafted feature methods (SCS and NFST)
rather than deep models. This is in contrast to the tests on CUHK01,
CUHK03, and Market-1501. 
This is due to (i) the small training data insufficient for learning
effectively discriminative deep models with millions of parameters;
(ii) the greater disparity to CUHK03 in camera viewing conditions
which makes knowledge transfer less effective (see Implementation).
% Such challenges similarly apply to the proposed JLML model. 
Nevertheless, the JLML model remains the best among all deep
methods with or without deep verification metric learning.
This validates the superiority and robustness of our deep joint global
and local representation learning of multi-loss classification
given sparse training data.
We attribute this property to the JLML's capability of 
mining complementary features in different context for both 
handling local misalignment and optimising global matching.

\begin{table} [!h]
	\centering
	\footnotesize
	\renewcommand{\arraystretch}{1}
	\setlength{\tabcolsep}{0.35 cm}
	\vspace{-.3cm}
	\caption{\footnotesize
		VIPeR evaluation. $1^\text{st}/2^\text{nd}$ best in red/blue.
	}
	\vskip 0pt %\vskip -6pt
	\begin{tabular}{|c||c|cccc|}
		\hline
		%		Dataset  & 
		%		\multicolumn{4}{c|}{VIPeR} \\ \hline
		%		Type &\multicolumn{4}{c|}{316 test IDs} \\ \hline
		\multirow{1}{*}{Cat}
		& Rank (\%) & R1 & R5 & R10 & R20 \\ \hline \hline
		\multirow{4}{*}{A}
		& XQDA 	& 40.0  & 68.1  &  80.5 &  91.1 \\   %\hline
		& GOG 	& 49.7  & -  &  88.7 &  94.5 \\  %\hline
		%
		% DNS 	& 42.3 & 71.5 & 82.9 & 92.1 \\
		& NFST 	&  \color{blue}  51.1  &  \color{blue}  82.1  &  \color{blue}  90.5 &  \color{blue}  95.9 \\
		& SCS 	& \color{red} 53.5  & \color{red} 82.6  & \color{red} 91.5 & \color{red} 96.7 \\ 
		\hline
		\multirow{2}{*}{B}
		& DCNN+ & 34.8  & 63.6  &  75.6 &  84.5 \\  %\hline	
		% & SICI & 35.8 & - & - & - \\
		& MTDnet & 47.5 & 73.1 & 82.6 & - \\
		\hline
		\multirow{3}{*}{C}
		& MCP & 47.8  & 74.7  & 84.8 &  91.1 \\
		%\hline
		& DGD & 38.6  & -  &  - &  -\\ 
		& {\bf JLML}  & 50.2  & 74.2 & 84.3 & 91.6 \\
		% {\bf JLML}(fusion)  & ?  & ? & ? & ? \\ 
		\hline
	\end{tabular}%}
	\label{tab:res_viper}
	\vspace{-.3cm}
\end{table}

\vspace{0.1cm}
\noindent {\bf (V) Evaluation on GRID. }
We compared JLML against $4$ competing methods on GRID\footnote{The
GRID dataset has not been evaluated as extensively as others like
VIPeR / CUHK01 / CUHK03, although GRID provides a more realistic test
setting with
a large number of distractors in testing. 
%One possible reason may be the much higher re-id matching challenge,
%as signified by the significantly lower matching rates in Table \ref{tab:res_grid}.
One possible reason is the more challenging re-id setting imposed by GRID resulting in significantly poorer matching rates by all published methods
(see \url{http://personal.ie.cuhk.edu.hk/~ccloy/downloads_qmul_underground_reid.html}), 
also as verified by our evaluation in Table \ref{tab:res_grid}.
}.
In addition to poor image resolution, poor lighting and 
a small training size ($125$ people),
GRID also has extra distractors in the testing population therefore
presenting a very challenging but realistic re-id scenario.
Table \ref{tab:res_grid} shows a significant superiority of JLML 
over existing state-of-the-arts, with Rank-1 $12.8\%$(37.5-24.7)
better than the 2nd best method GOG, a $51.8\%$ relative improvement.
This demonstrates the unique and practically desirable advantage of JLML in handling
more realistically challenging open-world re-id matching where
large numbers of distractors are usually present.
It is worth pointing out that this step-change advantage in re-id matching rate on GRID
is achieved by deep learning from only a limited number of training
identity classes with highly imbalanced images sampled from $8$ distributed camera views,
%This exception may be due to the per-camera image imbalance problem on this dataset, e.g., 
e.g. $25$ images from the $6^\text{th}$ camera vs. 
$513$ from the $5^\text{th}$ camera.
This imbalanced sampling directly results in 
not only scarce pairwise training data but also
insufficient training samples for pairwise camera views, resulting in
significant degradation in re-id performance from {\em all} pairwise
supervised learning based models XQDA, GOG, SCS, and X-Corr. 
In contrast, JLML is designed to avoid the need for pairwise labelled
information in model learning by instead learning from multi-loss classifications.
Moreover, the joint learning of multi-loss classification benefits
from concurrent local and global feature selections in different
context, resulting in more robust and accurate re-id matching in a
heterogeneous search space.

\begin{table} [!h]
	\centering
	\footnotesize
	\renewcommand{\arraystretch}{1}
	\setlength{\tabcolsep}{0.35 cm}
	\vspace{-.5cm}
	\caption{\footnotesize
		GRID evaluation.
		$1^\text{st}/2^\text{nd}$ best in red/blue.
	}
	\vskip 0pt %\vskip -6pt
	\begin{tabular}{|c||c|cccc|}
		\hline
		%		Dataset  & 
		%		\multicolumn{4}{c|}{VIPeR} \\ \hline
		%		Type &\multicolumn{4}{c|}{316 test IDs} \\ \hline
		\multirow{1}{*}{Cat}
		& Rank (\%) & R1 & R5 & R10 & R20 \\ \hline \hline
		\multirow{3}{*}{A}
		& XQDA 	& 16.6 & 33.8 & 41.8 & 52.4  \\   %\hline
		& GOG 	&  \color{blue}  24.7 &  \color{blue}  47.0 &  \color{blue}  58.4 &  \color{blue}  69.0 \\  %\hline
		%
		% 		DNS 	& \\
		%	DNS(fusion)	&  \\
		& SCS 	& 24.2 & 44.6 & 54.1 & 65.2 \\ 
		% \hline
		% DCNN+ 	& \\  %\hline	
		% MCP     & \\  %\hline
		% DGD    &  \\ 
		% MTDnet &  \\
		\hline
		\multirow{1}{*}{B}     
		& {X-Corr}  & 19.2  & 38.4 & 53.6 & 66.4 \\
		\hline
		\multirow{1}{*}{C}
		& {\bf JLML}  &   \color{red} 37.5 &  \color{red} 61.4  &  \color{red} 69.4  &   \color{red} 77.4  \\
		% {\bf JLML}(fusion)  & ?  & ? & ? & ? \\ 
		\hline
	\end{tabular}%}
	\label{tab:res_grid}
	\vspace{-.3cm}
\end{table}

\subsection{CNN Architecture Comparisons}
We compared the proposed JLML-ResNet39 model with 
four seminal classification CNN architectures
(Alexnet \cite{krizhevsky2012imagenet},
VGG16 \cite{simonyan2014very},
GoogLeNet \cite{szegedy2015going}, and
ResNet50 \cite{he2016deep}) in model size and complexity.
Table \ref{tab:base_nets} shows that 
the JLML has both the $2^\text{nd}$ smallest model size (7.2 million parameters) 
and the $2^\text{nd}$ smallest FLOPs ($1.54\!\times\! 10^9$),
although containing more streams (5 vs. 1 in all other CNNs)
and more layers ($39$, more than all except ResNet50). 

\begin{table} [!h]
	\centering
	% \scriptsize
	\footnotesize
	%\scalebox{0.8}{
	\renewcommand{\arraystretch}{1}
	\setlength{\tabcolsep}{0.12 cm}
	\vspace{-0.3cm}
	\caption{ \footnotesize
		Comparisons of model size and complexity.
		FLOPs: the number of FLoating-point OPerations;
		PN: Parameter Number.
	}
	\vskip 0pt %\vskip -6pt
	\begin{tabular}{|c||c|c|c|c|}
		\hline
		%		Model 
		%		& AlexNet & VGG16 & ResNet50 & GoogLeNet & \bf JLML-ResNet39 \\ \hline \hline
		%		FLOPs & \bf 7.25$\times$10$^8$ & 1.55$\times$10$^{10}$ & 3.80$\times$10$^9$ & 1.57$\times$10$^9$ & 1.54$\times$10$^9$  \\ % & 239.8\\   
		%		\hline
		%		PN (m) & 58.3 & 134.2  & 23.5 & \bf 6.0 & 7.2 \\
		%
		%
		Model & FLOPs & PN (million) & Depth & Stream \#
		\\ \hline
		AlexNet &  \bf 7.25$\times$10$^8$ & 58.3 & 7 & 1 \\ % & 239.8\\   %\hline
		VGG16 & 1.55$\times$10$^{10}$ & 134.2 & 16 & 1 \\ % & 549.4 \\   %\hline
		ResNet50 & 3.80$\times$10$^9$ & 23.5 & \bf 50 & 1 \\ % & 100.4 \\   
		GoogLeNet & 1.57$\times$10$^9$ & \bf 6.0 & 22 & 1 \\ \hline % & 50.5 \\ \hline
		{\bf JLML-ResNet39} & 1.54$\times$10$^9$ & 7.2 & 39 & \bf 5 \\ % & \bf 34.6\\ 
		\hline
	\end{tabular}%}
	\label{tab:base_nets}
	\vspace{-0.3cm}
\end{table}
%\begin{table} [!h]
%	\centering
%	\footnotesize
%	%\scalebox{0.8}{
%	\renewcommand{\arraystretch}{1}
%	\setlength{\tabcolsep}{0.10 cm}
%	\vspace{-0.3cm}
%	\caption{\footnotesize
%		Comparisons on Market-1501 using different CNN architectures.
%		FLOPs: the number of FLoating-point OPerations;
%		PN(m): Parameter numbers in million.
%	}
%	\vskip 0pt %\vskip -6pt
%	\begin{tabular}{|c|cc|cc|c|c|}
%		\hline
%		%		Dataset  & 
%		%		\multicolumn{4}{c|}{Market-1501} \\ \hline % \citep{zheng2015scalable}
%		Query Type &  \multicolumn{2}{c|}{Single-Query} & \multicolumn{2}{c|}{Multi-Query} & \multirow{2}{*}{FLOPs} & {PN} \\ \cline{1-5}%\hline
%		Measure (\%) 
%		& R1 & mAP & R1 & mAP  & & (m) \\ \hline \hline
%		AlexNet &  53.9 &  30.1  &  62.9 &  38.0 & \bf 7.25$\times$10$^8$ & 58.3 \\ % & 239.8\\   %\hline
%		VGG16 & 64.6  &  37.9  & 73.8 &  53.1  & 1.55$\times$10$^{10}$ & 134.2 \\ % & 549.4 \\   %\hline
%		ResNet50 &  74.2 &  51.9  &  78.2 & 57.7 & 3.80$\times$10$^9$ & 23.5 \\ % & 100.4 \\   
%		GoogLeNet & 76.9 & 52.5  &  83.3 &  63.4& 1.57$\times$10$^9$ & \bf 6.0 \\ \hline % & 50.5 \\ \hline
%		{\bf JLML-ResNet39} &  \textbf{85.1} &  \textbf{65.5}  &  \textbf{89.7} &  \textbf{74.5}&1.54$\times$10$^9$ & 7.2 \\ % & \bf 34.6\\ 
%		\hline
%	\end{tabular}%}
%	\label{tab:base_nets}
%	\vspace{-0.3cm}
%\end{table}
%======================= For future use ====================================
%======================= For future use ====================================
%======================= For future use ====================================

\subsection{Further Analysis and Discussions}
\label{sec:ablation}
We further examined the component effects of our JLML model
on Market-1501 in the following aspects.

\vspace{0.1cm}
\noindent {\bf (I) Complementary Benefits of Global and Local Features.}
We evaluated the complementary effects of our jointly learned local and global features
by comparing their individual re-id performance against that of the joint features.
Table \ref{tab:complementary} shows: 
{\bf (i)} Any of the two feature representations {\em alone} is
competitive for re-id, e.g. the local JLML feature surpasses S-CNN (Table \ref{tab:res_market}) 
by Rank-1 $13.1\%$(78.9-65.8) (SQ) and $10.4\%$(86.4-76.0) (MQ);
and by mAP $18.3\%$(57.8-39.5) (SQ) and $20.0\%$(68.4-48.4) (MQ). 
{\bf (ii)} A further performance gain is obtained from the joint
feature representation, yielding further $6.2\%$(85.1-78.9) (SQ) and
$3.3\%$(89.7-86.4) (MQ) in Rank-1 increase, and $7.7\%$(65.5-57.8)
(SQ) and $6.1\%$(74.5-68.4) (MQ) in mAP boost.
These results show the complementary advantages of jointly learning
the local and global features in different context using the JLML
model. 
%This shall be due to the good balancing between inter-branch correlation and independence
%relationships in our joint feature learning model (Sec. \ref{sec:JLML_model}).
%
%
% TO BE ADDED FOR JOURNAL VERSION
%We provide in Figure \ref{fig:show_complementary} a qualitative evaluation on
%the complementary effects between the global and local features by visualising 
%the appearance patterns learned concurrently by conv XX/XX/XX layers from the corresponding branches.
 
\begin{table} [!h]
	\centering
	\footnotesize
	%\scalebox{0.8}{
	\renewcommand{\arraystretch}{1}
	\setlength{\tabcolsep}{0.4 cm}
	\vspace{-0.3cm}
	\caption{\footnotesize
		Complementary benefits of global and local features.
	}
	\vskip 0pt %\vskip -6pt
	\begin{tabular}{|c||cc|cc|}
		\hline
		%		Dataset  & 
		%		\multicolumn{4}{c|}{Market-1501} \\ \hline % \citep{zheng2015scalable}
		Query Type &  \multicolumn{2}{c|}{Single-Query} &\multicolumn{2}{c|}{Multi-Query} \\  \cline{1-5}% \hline
		Measure (\%)    
		& R1 & mAP & R1 & mAP  \\ \hline \hline

		JLML (Global) & 77.4 & 56.0 & 85.0 & 66.0\\   %\hline
		JLML (Local) &  78.9& 57.8 & 86.4 & 68.4\\ \hline
		JLML (joint) & \textbf{85.1} &  \textbf{65.5}  &  \textbf{89.7} &  \textbf{74.5} \\
		\hline
	\end{tabular}%}
	\label{tab:complementary}
	\vspace{-0.3cm}
\end{table}

% ADD FOR JOURNAL VERSION
%\begin{figure} [ht]
%	\centering
%	% \includegraphics[width=1.0\linewidth]{images/datasets.pdf}
%	\vskip -0.5cm
%	\caption{\footnotesize
%		Qualitative evaluation on the complementary effects between
%		the global and local feature representations induced by the proposed JLML model.
%		({\color{red} TODO})
%	}
%	\label{fig:show_complementary}
%\end{figure}

\vspace{0.1cm}
\noindent {\bf (II) Importance of Branch Independence.}
We evaluated the importance of branch independence by comparing our {\em MultiLoss} design
with a {\em UniLoss} design that merges 
two branches into a single loss \cite{Cheng_TCP}.
%({\color{red}TODO: More details may be needed here})
Table \ref{tab:individuality} shows that the proposed MultiLoss model significantly
improves the discriminative power of global and local re-id features, e.g. with 
Rank-1 increase of $9.0\%$(85.1-76.1) (SQ) and $6.0\%$(89.7-83.7) (MQ);
and mAP improvement of $13.3\%$(65.5-52.2) (SQ) and $11.7\%$(74.5-62.8) (MQ).
This shows that branch independence plays a critical role in 
joint learning of multi-loss classification for effective feature optimisation.
One plausible reason is due to the negative effect of a single loss imposed
on the learning behaviour of both branches, caused by the potential divergence
in discriminative features in different context (local and global).
This is shown by the significant performance degradation of both global and local features
when the UniLoss model is imposed. 

\begin{table} [!h]
	\centering
	\footnotesize
	%\scalebox{0.8}{
	\renewcommand{\arraystretch}{1}
	\setlength{\tabcolsep}{0.2 cm}
	\vspace{-0.3cm}
	\caption{\footnotesize
		Importance of branch independence.
	}
	\vskip 0pt %\vskip -6pt
	\begin{tabular}{|c|c||cc|cc|}
		\hline
		%		Dataset  & 
		%		\multicolumn{4}{c|}{Market-1501} \\ \hline % \citep{zheng2015scalable}
		\multirow{2}{*}{Loss}
		& Query Type &  \multicolumn{2}{c|}{Single-Query} &\multicolumn{2}{c|}{Multi-Query} \\  \cline{2-6}% \hline
		& Measure (\%)    
		& R1 & mAP & R1 & mAP  \\ \hline \hline
		\multirow{3}{*}{UniLoss} 
		& Global Feature & 58.3 & 31.7 & 70.4 & 43.2\\   %\hline
		& Local Feature &  46.3 & 26.3 & 58.0 & 34.0\\ \cline{2-6}
		& \bf Full & 76.1 & 52.2 & 83.7 & 62.8  \\ 
		\hline   %\hline
		\multirow{3}{*}{\bf MultiLoss} 
		& Global Feature & 77.4 & 56.0 & 85.0 & 66.0\\   %\hline
		& Local Feature &  78.9& 57.8 & 86.4 & 68.4\\ \cline{2-6}
		& {\bf Full} & \textbf{85.1} &  \textbf{65.5}  &  \textbf{89.7} &  \textbf{74.5} \\
		\hline
	\end{tabular}%}
	\label{tab:individuality}
	\vspace{-0.3cm}
\end{table}

\vspace{0.1cm}
\noindent {\bf (III) Benefits from Shared Low-Level Features.}
We evaluated the effects of interaction between global and local branches
introduced by the shared conv layer (common ground) by deliberately 
removing it and then comparing the re-id performance.
Table \ref{tab:common_conv} shows the benefits from 
jointly learning low-level features in the common conv layers,
e.g. improving Rank-1 by $1.9\%$(85.1-83.2) / $1.4\%$(89.7-88.3)
and mAP by $2.4\%$(65.5-63.1) / $2.4\%$(74.5-72.1) 
for single-/multi-query re-id.
This confirms a similar finding as in multi-task learning study \cite{argyriou2007multi}.

\begin{table} [!h]
	\centering
	\footnotesize
	%\scalebox{0.8}{
	\renewcommand{\arraystretch}{1}
	\setlength{\tabcolsep}{0.3 cm}
	\vspace{-0.3cm}
	\caption{\footnotesize
		Benefits from shared low-level features.
	}
	\vskip 0pt %\vskip -6pt
	\begin{tabular}{|c|cc|cc|}
		\hline
		%		Dataset  & 
		%		\multicolumn{4}{c|}{Market-1501} \\ \hline % \citep{zheng2015scalable}
		Query Type &  \multicolumn{2}{c|}{Single-Query} &\multicolumn{2}{c|}{Multi-Query} \\ \hline
		Measure (\%)    
		& R1 & mAP & R1 & mAP  \\ \hline \hline
		{\bf Without} Shared Feature & 83.2 & 63.1 & 88.3 & 72.1
		\\ \hline
		{\bf With} Shared Feature & 
		\textbf{85.1} &  \textbf{65.5}  &  \textbf{89.7} &  \textbf{74.5} \\ \hline   %\hline
	\end{tabular}%}
	\label{tab:common_conv}
	\vspace{-0.3cm}
\end{table}

\vspace{0.1cm}
\noindent {\bf (IV) Effects of Selective Feature Learning. }
We evaluated the contribution of our structured sparsity 
based Selective Feature Learning (SFL)
(Eq. \eqref{eq:loss}).
Table \ref{tab:selective_learn} shows that our SFL mechanism can
bring additional re-id matching benefits, e.g. improving Rank-1 rate
by $1.7\%$(85.1-83.4) (SQ) and $1.0\%$(89.7-88.7) (MQ); and mAP by
$1.7\%$(65.5-63.8) (SQ) and $1.6\%$(74.5-72.9) (MQ). 

\begin{table} [!h]
	\centering
	\footnotesize
	%\scalebox{0.8}{
	\renewcommand{\arraystretch}{1}
	\setlength{\tabcolsep}{0.4 cm}
	\vspace{-0.3cm}
	\caption{\footnotesize
		Effects of selective feature learning (SFL).
	}
	\vskip 0pt %\vskip -6pt
	\begin{tabular}{|c|cc|cc|}
		\hline
		%		Dataset  & 
		%		\multicolumn{4}{c|}{Market-1501} \\ \hline % \citep{zheng2015scalable}
		Query Type &  \multicolumn{2}{c|}{Single-Query} &\multicolumn{2}{c|}{Multi-Query} \\ \hline
		Measure (\%)    
		& R1 & mAP & R1 & mAP  \\ \hline \hline
		Without SFL &  83.4 &  63.8  &  88.7 &  72.9\\ \hline
		{\bf With SFL} &  \textbf{85.1} &  \textbf{65.5}  &  \textbf{89.7} &  \textbf{74.5}\\ 
		\hline
	\end{tabular}%}
	\label{tab:selective_learn}
	\vspace{-0.3cm}
\end{table}

\vspace{0.1cm}
\noindent {\bf (V) Choice of Generic Matching Metrics.}
We evaluated the choice of generic matching distances on person re-id
using the full JLML feature.
Table \ref{tab:match_dist} shows that L1 and L2 generate very similar and competitive re-id matching accuracies.
This suggests the flexibility of the JLML model in adopting generic matching metrics.

\begin{table} [!h]
	\centering
	\footnotesize
	%\scalebox{0.8}{
	\renewcommand{\arraystretch}{1}
	\setlength{\tabcolsep}{0.4 cm}
	\vspace{-0.3cm}
	\caption{\footnotesize
		Effects of generic matching metrics.
	}
	\vskip 0pt %\vskip -6pt
	\begin{tabular}{|c|cc|cc|}
		\hline
		%		Dataset  & 
		%		\multicolumn{4}{c|}{Market-1501} \\ \hline % \citep{zheng2015scalable}
		Query-Type &  \multicolumn{2}{c|}{Single-Query} &\multicolumn{2}{c|}{Multi-Query} \\ \hline
		Measure (\%)    
		& R1 & mAP & R1 & mAP  \\ \hline \hline
		% Cosine & 85.1 & 65.5 & 89.7 & 74.5\\ \hline
		L1 & 84.9 & 65.3 & 89.2 & 74.6 \\ \hline
		L2 &  {85.1} &  {65.5}  &  {89.7} & {74.5}\\ 
		\hline
	\end{tabular}%}
	\label{tab:match_dist}
	\vspace{-0.1cm}
\end{table}

\begin{figure}[h]
	\centering
    \includegraphics[width=1.0\linewidth]{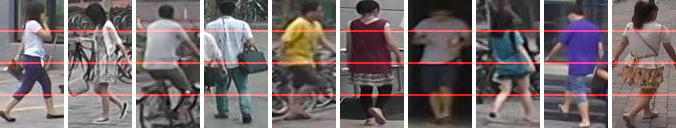}
%	\subfigure[\scriptsize VIPeR]{
%		\includegraphics[width=0.17\linewidth]{images/dataset_viper.png}
%	}
	\vskip -0.3cm
	\caption{\footnotesize 
		Visualisation of the optimal body part decomposition.
	}
	\label{fig:local_parts}
\end{figure}

%\begin{table} [!h]
%	\centering
%	\footnotesize
%	%\scalebox{0.8}{
%	\renewcommand{\arraystretch}{1}
%	\setlength{\tabcolsep}{0.1 cm}
%	\vspace{-0.3cm}
%	\caption{\footnotesize
%		Effects of selective learning.
%	}
%	\vskip 0pt %\vskip -6pt
%	\begin{tabular}{|c|cc|cc|}
%		\hline
%%		Dataset  & 
%%		\multicolumn{4}{c|}{Market-1501} \\ \hline % \citep{zheng2015scalable}
%		Query Type &  \multicolumn{2}{c|}{Single-Query} &\multicolumn{2}{c|}{Multi-Query} \\ \hline
%		Measure (\%)    
%		& R1 & mAP & R1 & mAP  \\ \hline \hline
%		global\{$\|\textbf{w}_G\|_2 $\} (a)&  74.3 & 52.8  &  82.4 &  62.9\\   %\hline
%		local\{$\|\textbf{w}_L\|_2 $\} (b)&  79.5 & 58.4  &  86.4 &  68.4\\
%		
%		global\{$\|\textbf{w}_G\|_2  + \|\textbf{w}_G\|_{2,1} $\} (c)&  77.4 &  56.0  &  85.0 &  66.0\\   %\hline
%		local\{$\|\textbf{w}_L\|_2  + \|\textbf{w}_L\|_{1,2} $\} (d)&  78.9 &  57.8  &  86.4 &  68.4\\ \hline   %\hline
%		a+b &  83.4 &  63.8  &  88.7 &  72.9\\ 
%		b+c &  84.4 &  65.3  &  89.1 &  73.4 \\
%		c+d &  \textbf{85.1} &  \textbf{65.5}  &  \textbf{89.7} &  \textbf{74.5}\\ 
%		\hline
%	\end{tabular}%}
%	\label{tab:res_deepmodels}
%	\vspace{-0.3cm}
%\end{table}
\noindent {\bf (VI) Effects of Body Parts Number.}
{We evaluated the sensitivity of local decomposition,
	i.e. body parts number $m$. 
	Table \ref{tab:num_body_part} shows that the decomposition of 4 body-parts is the optimal choice, approximately corresponding to head+shoulder, upper-body, upper-leg and lower-leg (Figure \ref{fig:local_parts}).}

\begin{table} [h]
	\centering
	\footnotesize
	%\scalebox{0.8}{
	\renewcommand{\arraystretch}{1}
	\setlength{\tabcolsep}{0.4 cm}
	\vspace{-0.3cm}
	\caption{\footnotesize
		Effects of body parts number.
	}
	\vskip 0pt %\vskip -6pt
	\begin{tabular}{|c|cc|cc|}
		\hline
		%		Dataset  & 
		%		\multicolumn{4}{c|}{Market-1501} \\ \hline % \citep{zheng2015scalable}
		Query-Type &  \multicolumn{2}{c|}{Single-Query} &\multicolumn{2}{c|}{Multi-Query} \\ \hline
		Measure (\%)     
		& R1 & mAP & R1 & mAP  \\ \hline \hline
		2 & 83.9 & 64.4 & 88.8 & 72.9 \\ \hline
		4 & \bf 85.1 &  \bf 65.5  &  \bf 89.7 & \bf 74.5 \\\hline
		6 & 83.4 &  62.6  &  88.5 & 71.8 \\\hline
		8 &  82.3 &  61.3  &  87.4 & 70.7 \\\hline
		10 &  81.7 &  60.4  &  87.2 & 69.8 \\
		\hline
	\end{tabular}%}
	\label{tab:num_body_part}
	\vspace{-0.3cm}
\end{table}

\vspace{0.1cm}
\noindent {\bf (VII) Complementary Effects between JLML Deep Features
	and Supervised Metric Learning. }
{We evaluated the complementary effects of the JLML deep features
	and conventional supervised metric learning (XQDA \cite{liao2015person},
	KISSME \cite{KISSME_CVPR12}, and CRAFT \cite{chen2017person}). 
	Results from Table \ref{tab:metric_learning} show that:
	(1) Given strong deep learning features such as JLML, 
	additional distance metric learning does not benefit further from the same training data. 
	(2) Moreover, it may even suffer from some adversary effect.}

\begin{table} [h]
	\centering
	\footnotesize
	%\scalebox{0.8}{
	\renewcommand{\arraystretch}{1}
	\setlength{\tabcolsep}{0.4 cm}
	%\vspace{-0.3cm}
	\caption{\footnotesize
	Complementary of JLML features and metric learning.
	}
	\vskip 0pt %\vskip -6pt
	\begin{tabular}{|c|cc|cc|}
		\hline
		%		Dataset  & 
		%		\multicolumn{4}{c|}{Market-1501} \\ \hline % \citep{zheng2015scalable}
		Query-Type &  \multicolumn{2}{c|}{Single-Query} &\multicolumn{2}{c|}{Multi-Query} \\ \hline
		Measure (\%)     
		& R1 & mAP & R1 & mAP  \\ \hline \hline
		KISSME %\cite{KISSME_CVPR12} 
		& 82.1 & 61.4 & 87.5 & 70.2\\
		XQDA %\cite{liao2015person} 
		& 82.6 & 63.2 & 88.2 & 72.4\\
		CRAFT %\cite{chen2017person} 
		& 77.9 & 56.4 & - & -\\
		\hline
		L2 & \bf 85.1 &  \bf 65.5  &  \bf 89.7 & \bf 74.5 \\
		\hline
	\end{tabular}%}
	\label{tab:metric_learning}
	\vspace{-0.3cm}
\end{table}

\begin{figure} [h]
	\centering
	\includegraphics[width=1.0\linewidth]{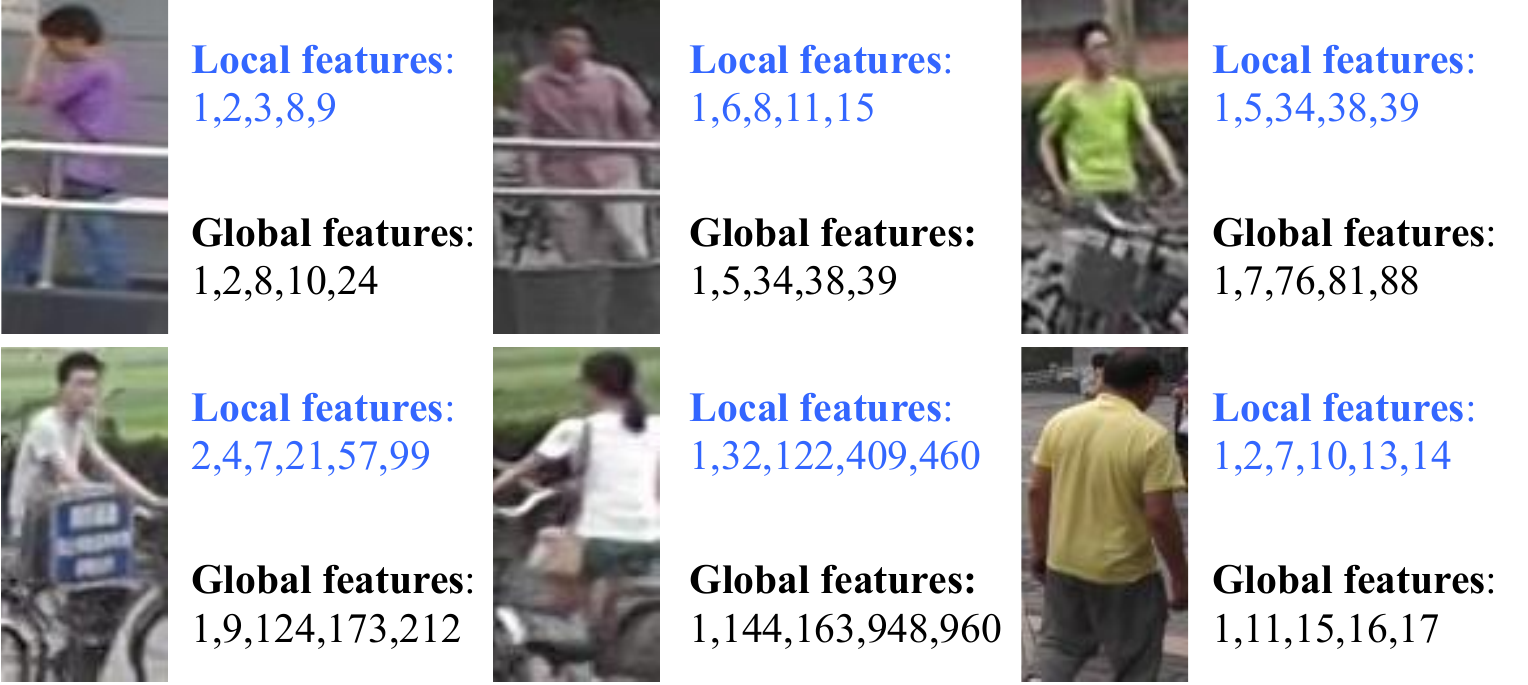}
	\vskip -0.3cm
	\caption{\footnotesize 
		% Examples of comparing local and global features.
		Comparing the gallery true match ranks of each probe image (single-query) with occlusion and/or misalignment by the local and global features. Each probe may have multiple truth matches in the gallery. Smaller numbers mean better ranking performances.
	}
	\label{fig:local_global}
\end{figure}

\vspace{0.1cm}
\noindent {\bf (VIII) Local Features vs. Global Features.}
{A strength of the local features is the capability 
	of mitigating misalignment and occlusion, as compared to the global features.  
	This is inherently learned from data by the JLML local branch. 
	Figure \ref{fig:local_global} shows the single-query re-id results on six randomly selected probe persons with misalignment and/or occlusion.
	% and compare the re-id ranks of gallery truth matches of these probes by local and global features respectively. 
	It is evident that the local features achieve better re-id matching ranks than the global counterparts in most cases.
	This clearly demonstrates the robustness of local features against the misalignment of and occlusion within a person bounding box.}

\vspace{0.1cm}
\noindent {\bf (IX) Feature Extraction Time Cost. }
{The average time for extracting JLML feature
	is 2.75 milliseconds per image (364 images per second)
	on a Nvidia Pascal P100 GPU card.}

%================================ 
\section{Conclusion}
In this work, we presented a novel Joint Learning of Multi-Loss (JLML)
CNN model (JLML-ResNet39)
for person re-identification feature learning.
In contrast to existing re-id approaches
that often employ either global or local appearance features alone,
the proposed model is capable of extracting and exploiting both and maximising their correlated complementary effects by
learning discriminative feature representations in different context subject to
multi-loss classification objectives in a unified framework.
This is made possible by the proposed JLML-ResNet39 architecture design.
Moreover, we introduce a structured sparsity based feature selective learning mechanism
to reduce feature redundancy and further improve the joint feature selections. 
Extensive comparative
evaluations on five re-id benchmark datasets 
were conducted to validate the advantages of
the proposed JLML model over a wide range of the state-of-the-art methods 
on both manually labelled and more challenging auto-detected person images.
We also provided component evaluations and analysis of model
performance in order to give insights on the model design.

% \section*{Acknowledgments}

%\appendix

\small
\bibliographystyle{named}
\bibliography{ijcai17_reid.bib}

\end{document}